# MOTOR COMPOUND FAULT DIAGNOSIS USING AN ATTENTION-BASED CROSS-TALK ARCHITECTURE

Wonjun Yi, Wonho Jung, Junho Kim, Rismaya Kumar Mishra and Yong-Hwa Park
*Korea Advanced Institute of Science and Technology (KAIST), Daejeon, Republic of Korea*
*email: lasscap@kaist.ac.kr*

As physical AI increasingly relies on sensing to perceive and understand physical systems, reliable condition monitoring of industrial machinery has become more important. Motors are ubiquitous in industrial systems, and increasing system complexity raises the likelihood of simultaneous faults. Accordingly, compound fault diagnosis that identifies multiple fault components and their severities is increasingly required. Because different fault components jointly affect the same vibration signal, their severity classification can be formulated as related tasks, making multi-task learning (MTL) suitable for exploiting shared information through inductive transfer. However, conventional shared-trunk architectures force tasks to share a common representation, potentially causing negative transfer because individual fault components are characterized by different time-frequency regions. To address this issue, this study proposes a Residual Neural Dimension Reductor with Time-Frequency Channel Attention (RNDR-TFCA), an attention-enhanced cross-talk architecture for multi-output compound fault diagnosis. RNDR-TFCA preserves task-specific feature extractors while selectively exchanging information across tasks. Time-frequency attention emphasizes informative intra-task regions, while channel attention enhances useful inter-task information after nonlinear feature fusion. Experiments on a rotor-bearing system with inner-race fault, outer-race fault, misalignment, and unbalance showed that RNDR-TFCA achieved a macro F1 score of 0.898, outperforming single-task, multi-class, shared-trunk, and existing cross-talk baselines. Ablation experiments further confirmed the benefit of jointly applying time-frequency and channel attention. RNDR-TFCA required 713 K parameters and 170.42 MMACs, only slightly higher than RNDR. Channel-expanded baselines still underperformed RNDR-TFCA, confirming that the gain is not simply due to increased model capacity.

## 1. Introduction

As physical AI increasingly relies on sensing to perceive the states of physical systems, reliable condition monitoring of industrial machinery has become increasingly important. Motors are widely used in industrial systems, and increasing system complexity with multiple interacting components raises the possibility of system-level compound faults. Consequently, compound fault diagnosis, which simultaneously identifies multiple fault components and their severities, has become increasingly important [1, 2]. In rotating machinery, bearing faults, misalignment, and unbalance may coexist and jointly influence the measured vibration signal.

A compound fault can be described by multiple fault elements with different severity levels. Because these elements simultaneously affect the same measured signal, their severity classification can be formulated as related tasks. MTL is therefore suitable for compound fault diagnosis by jointly learning these tasks and exploiting shared information through inductive transfer [3]. Compared with MCC, which treats each combination of fault severities as a separate class, multi-output classification estimates each fault element independently and avoids the rapid growth of joint classes as the number of fault elements increases [4].

However, complete feature sharing can cause negative transfer when tasks require different representations. In vibration-based diagnosis, individual fault elements exhibit different physical signatures and informative time-frequency regions. Thus, conventional shared-trunk MTL architectures [5, 6], which use a common feature extractor for all tasks, may transfer irrelevant information across tasks and degrade performance [7]. Cross-talk architectures [8-11] mitigate this issue by maintaining task-specific feature extractors while selectively exchanging intermediate features. Representative methods include CS [8], CC [9], and NDDR-CNN [10]. Our previously proposed RNDR [11] further preserves main-task information through a residual connection while integrating multiple sub-task features through channel-wise concatenation and nonlinear transformation. Recent cross-task attention approaches [12, 13] suggest that explicitly emphasizing informative intra-task and inter-task features can further improve cross-task information exchange.

Based on these observations, this study proposes Residual Neural Dimension Reductor with Time-Frequency Channel Attention (RNDR-TFCA) for motor compound fault diagnosis. The proposed architecture introduces two complementary attention mechanisms into the cross-talk layer. First, time-frequency attention is applied to each sub-task feature before feature concatenation to emphasize informative regions within each task. Subsequently, channel attention based on a squeeze-and-excitation (SE) block [14] emphasizes informative inter-task feature components after nonlinear feature fusion. Thus, RNDR-TFCA aims to enhance both intra-task and inter-task information while retaining task-specific representations. The proposed method is evaluated on vibration signals from a rotor-bearing system involving inner-race fault, outer-race fault, misalignment, and unbalance under various operating conditions. In addition to comparisons with single-task learning (STL)-based classifiers, MCC, shared-trunk architectures [5, 6], and cross-talk architectures [8–11], an ablation study is conducted to evaluate the contribution of each attention mechanism in RNDR-TFCA. Computational complexity was further evaluated using the number of parameters and MACs, and tenfold channel-expanded baselines were additionally compared to determine whether the performance improvement was simply attributable to increased model capacity.

## 2. Multi-task learning

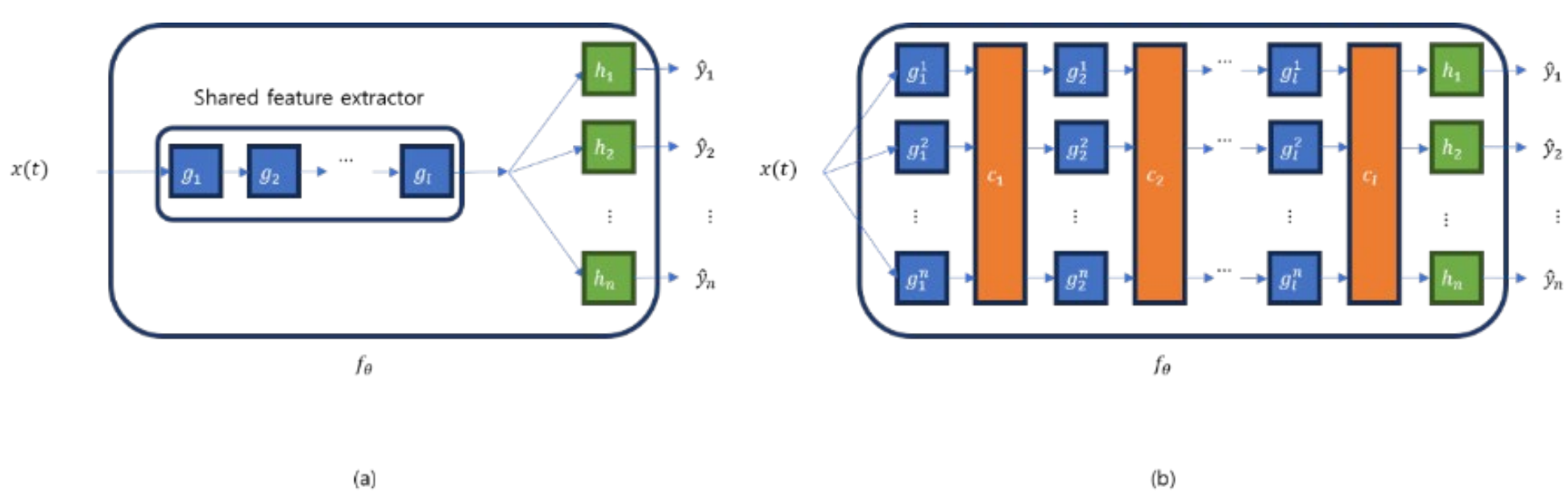

Figure 1: Architecture of multi-task learning models: (a) Shared trunk architecture, (b) Cross-talk architecture.

### 2.1 Shared trunk architecture

Shared-trunk architectures are multi-task learning models in which multiple tasks share a common feature extractor, referred to as the shared trunk, as illustrated in Fig. 1(a). In this study, the task-constrained deep convolutional network (TCDCN) [5] and the multi-task attention network (MTAN) [6] were adopted as representative shared-trunk models. TCDCN consists of a shared feature extractor followed by multiple task-specific classification branches. MTAN follows a similar overall structure but additionally incorporates task-specific attention modules to refine the features used by each task. A shared trunk can learn common representations across related tasks; however, forced feature sharing may cause negative transfer when task-specific feature requirements differ.

### 2.2 Cross-talk architecture

Cross-talk architectures employ separate task-specific feature extractors interconnected by CTLs, as conceptually illustrated in Fig. 1(b). Unlike shared-trunk architectures, each task can maintain its own feature representation while receiving complementary information from the other tasks through the CTLs. This selective feature exchange allows cross-talk architectures to exploit task relationships while reducing the risk of negative transfer caused by indiscriminate feature sharing.
In this study, CS [8], CC [9], NDDR-CNN [10], and RNDR [11] were considered as representative cross-talk architectures. CS exchanges task features through learnable linear combinations, while CC uses non-linear convolutional transformations. NDDR-CNN concatenates task features along the channel dimension and projects them back to the original dimension. RNDR [11] further preserves the original task-specific feature through a residual connection. However, RNDR does not explicitly emphasize informative intra-task regions or inter-task feature components, leaving room for more selective feature exchange.

### 2.3 Cross-task attention module

Recently, cross-task attention mechanisms (CTAMs) have been introduced for dense scene prediction to strengthen intra-task and inter-task features during information exchange [12, 13]. This concept can complement RNDR by emphasizing informative features within and across tasks. However, existing CTAMs were mainly designed for dense prediction and involve relatively complex feature-interaction modules. Therefore, a more compact attention mechanism tailored to vibration-based fault diagnosis is required. Based on this motivation, we propose RNDR-TFCA, which incorporates time-frequency and channel attention to enhance intra-task and inter-task features, respectively.

## 3. RNDR-TFCA

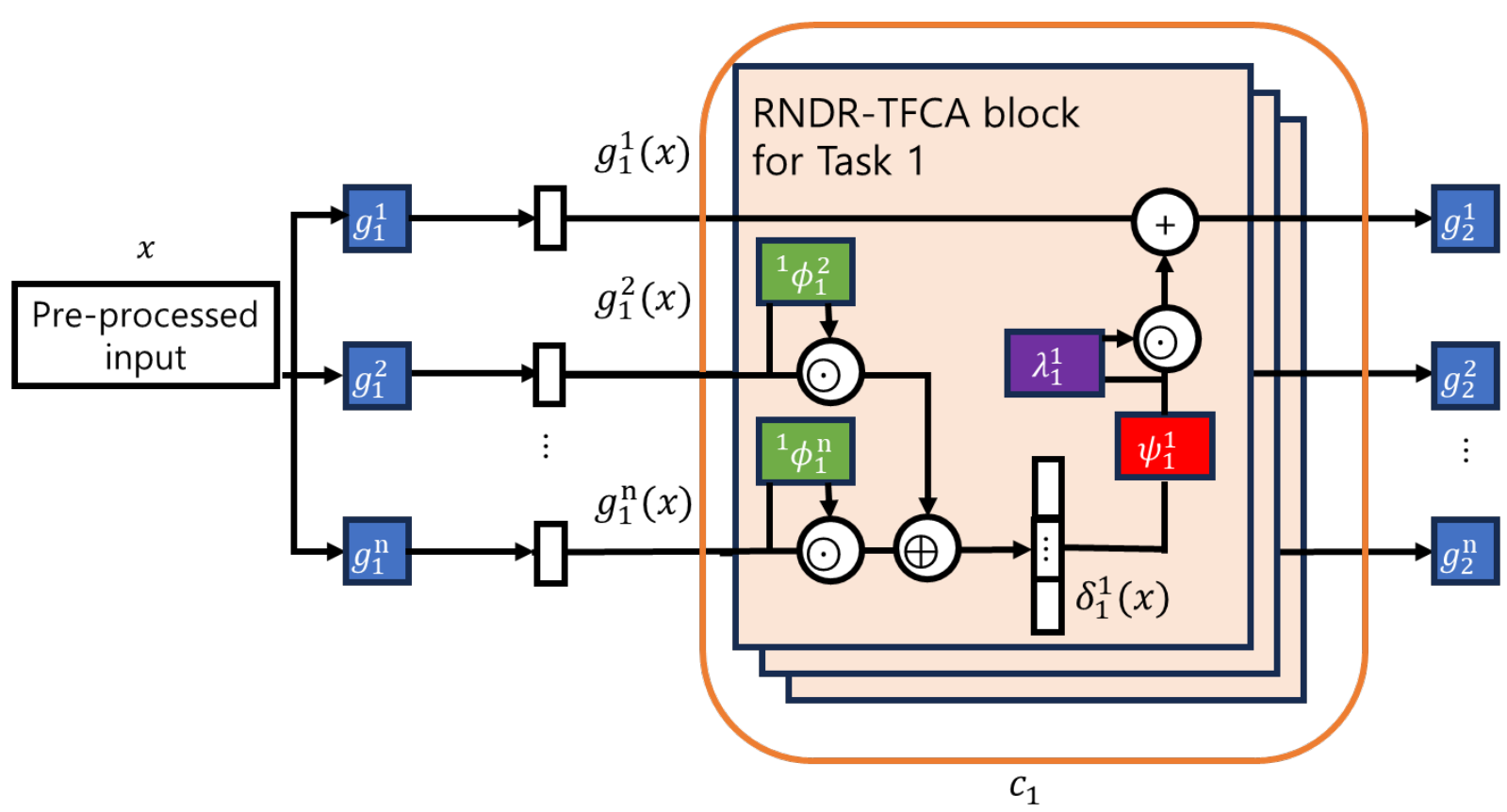

Figure 2: RNDR-TFCA block.

Fig. 2 illustrates the proposed RNDR-TFCA architecture. RNDR-TFCA retains the basic structure of RNDR [11], in which each task has an independent feature extractor and information from the other tasks is transferred through the CTL. The original main-task feature is preserved through a residual connection, while the sub-task features are combined and transformed before being added to the main-task path.
To enhance intra-task information, a time-frequency attention block is first applied independently to each sub-task feature before channel concatenation. This operation emphasizes informative regions in the time-frequency representation while suppressing less relevant regions. The attention-weighted sub-task features are then concatenated along the channel dimension and processed through a convolutional layer with nonlinear activation, following the basic feature-fusion strategy of RNDR.
After the fused feature is obtained, channel attention is applied to emphasize informative inter-task feature components before they are transferred to the main-task path. Specifically, an SE block [14] adaptively weights the channels of the fused representation according to their relevance. The resulting feature is finally combined with the preserved main-task feature through the residual connection. In this manner, RNDR-TFCA enhances both intra-task and inter-task information within the CTL while maintaining the task-specific feature extraction structure of RNDR.

# 4. Experiment setup

## 4.1 Dataset configuration

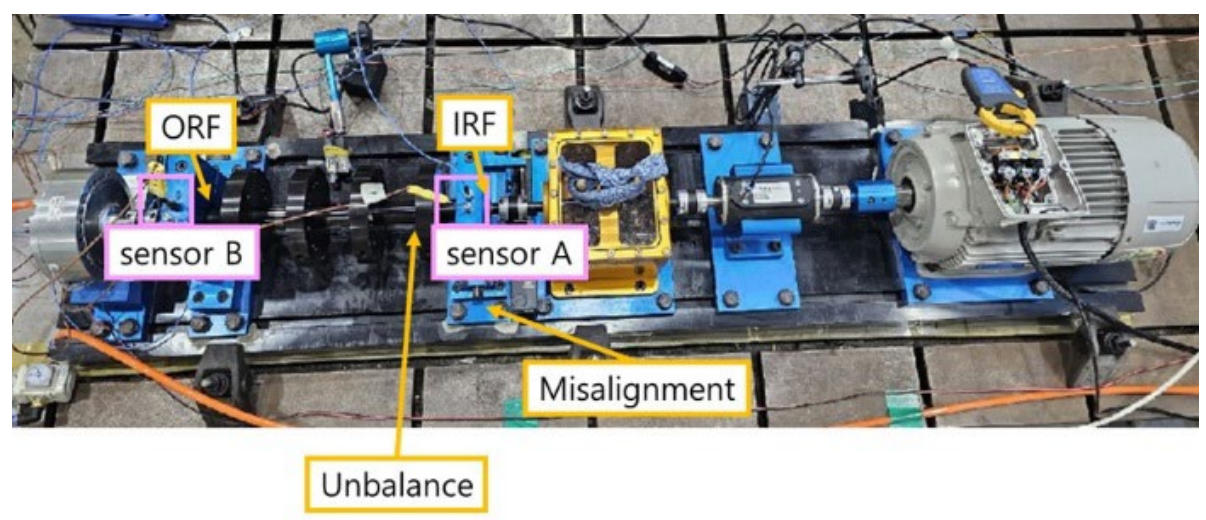

Figure 3: Motor test-bed used for data acquisition.

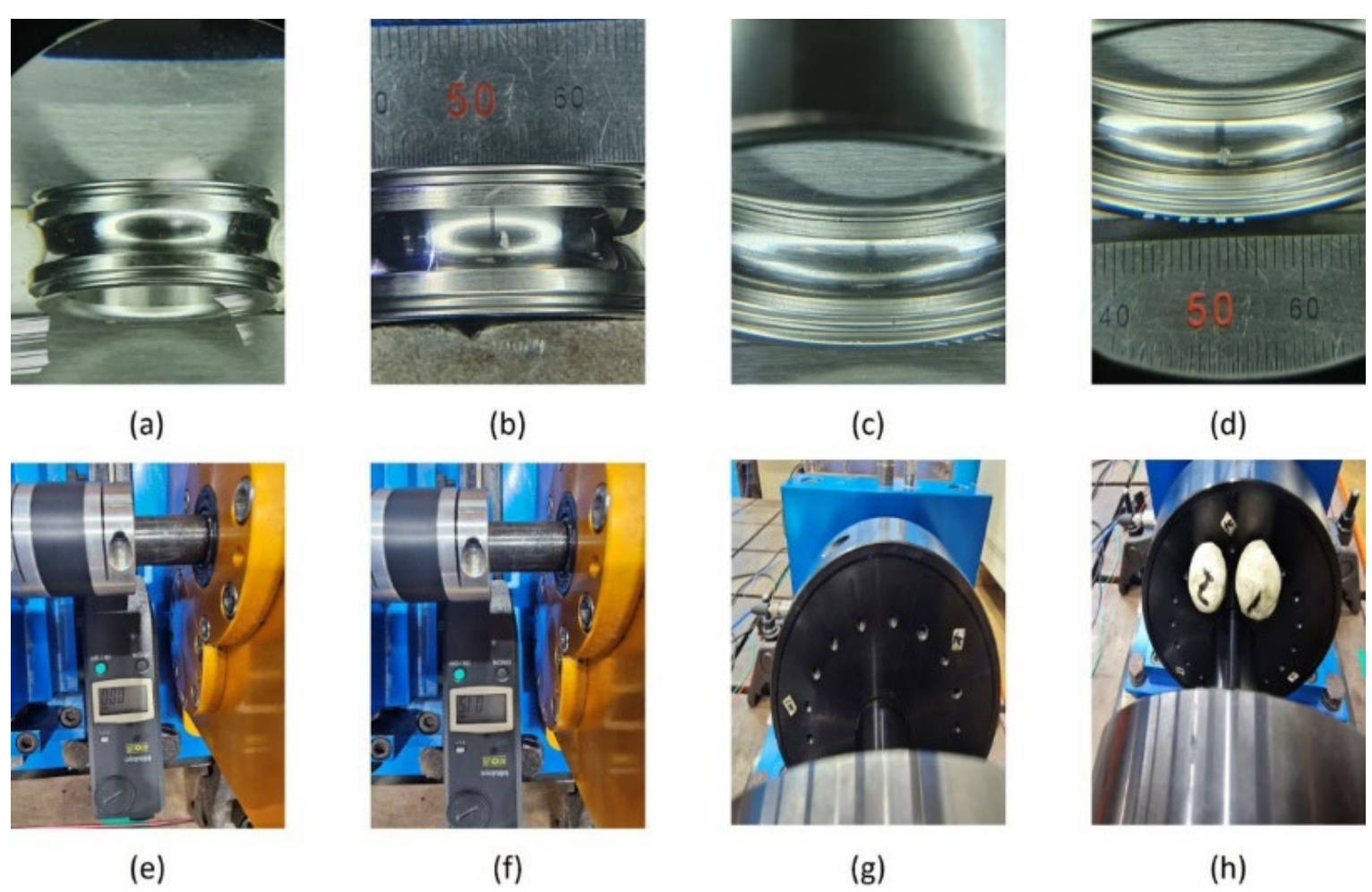

Figure 4: Each element with and without fault: (a) normal inner race, (b) cracked inner race, (c) normal outer race, (d) cracked outer race, (e) normal shaft, (f) misaligned shaft, (g) normal disk, (h) disk with unbalance

The experiments were conducted using the motor test-bed [11] shown in Fig. 3. Vibration signals were acquired from two measurement locations, denoted as sensor A and sensor B, while four fault components were independently introduced into the system: inner-race fault (IRF), outer-race fault (ORF), misalignment, and unbalance. The physical configurations of the normal and faulty elements are presented in Fig. 4.
Each fault component was assigned discrete severity levels. IRF and ORF were classified into two levels, corresponding to 0 and 0.2 mm, whereas misalignment was configured at 0, 0.15, and 0.3 mm. Unbalance was represented by three mass conditions of 0, 10, and 18 g. Consequently, each sample represents a compound fault condition defined by the combination of these four fault-element severities. The resulting dataset was class-imbalanced, with an overall imbalance ratio of 18.15.

A total of 70 min of vibration data was recorded at a sampling frequency of 25.6 kHz. The continuous signals were segmented into 4-s samples. To include variations in operating conditions, three rotational-speed patterns were employed: sinusoidal, triangular, and constant RPM. In this study, Subset E of the publicly available NOVIC+ motor compound fault dataset (https://zenodo.org/records/15743425) was used, which combines all three operating conditions. The resulting samples were divided into training, validation, and test sets with a ratio of 2:1:1.

### 4.2 Implementation and evaluation

Vibration signals were transformed into time-frequency representations using STFT before being fed into the networks. Unless otherwise specified, the feature extractor consisted of three convolutional layers, while each task-specific classifier consisted of two fully connected layers. The same basic backbone configuration was used across the compared architectures where applicable to provide a consistent comparison.

For the multi-output classification models, separate task-specific branches were used to estimate the severity of each fault component. In contrast, the MCC did not employ separate task-specific branches. Instead, all possible combinations of the four fault components were treated as individual classes, resulting in a single classifier with 36 output classes.

All models were trained for 100 epochs using the Adam optimizer [15] with an initial learning rate of $1 \times 10^{-3}$ and a batch size of 16. Categorical cross-entropy loss with label smoothing [16] was used for optimization. The validation loss was evaluated after each epoch, and the checkpoint yielding the lowest validation loss was selected for testing. Classification performance was evaluated using the Macro F1 score calculated over the overall motor compound fault states. The experiments were conducted on a system equipped with an NVIDIA GeForce RTX 4070 GPU and an AMD Ryzen 7 7700 8-Core Processor.

## 5. Result

### 5.1 Performance comparison

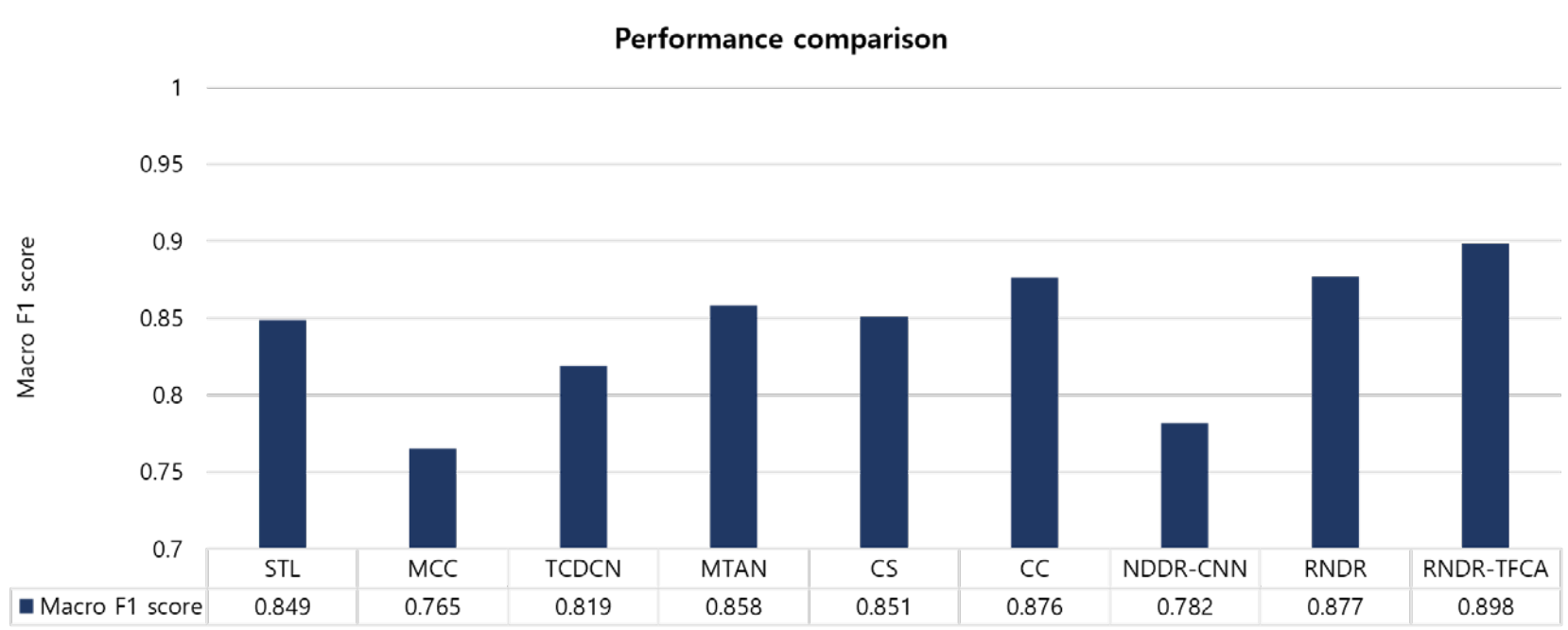

Figure 5: Comparison of Macro F1 scores among STL, MCC, MTL architectures, and proposed RNDR-TFCA.

Fig. 5 compares the macro F1 scores of the evaluated methods for motor compound fault diagnosis. Among the baseline methods, CC [9] achieved the highest performance with a macro F1 score of 0.876, followed by MTAN [6] with 0.858 and CS [8] with 0.851. The previously proposed RNDR [11] further improved the score to 0.877, indicating the effectiveness of its residual cross-talk structure.

The proposed RNDR-TFCA achieved the highest macro F1 score of 0.898, outperforming all compared methods. In particular, it improved the performance by 0.021 over RNDR and by 0.022 over CC, the best-performing conventional cross-talk baseline. These results indicate that incorporating time-frequency and channel attention into the RNDR-based CTL enables more effective intra-task feature enhancement and inter-task information exchange for compound fault diagnosis.

## 5.2 Ablation study of attention blocks

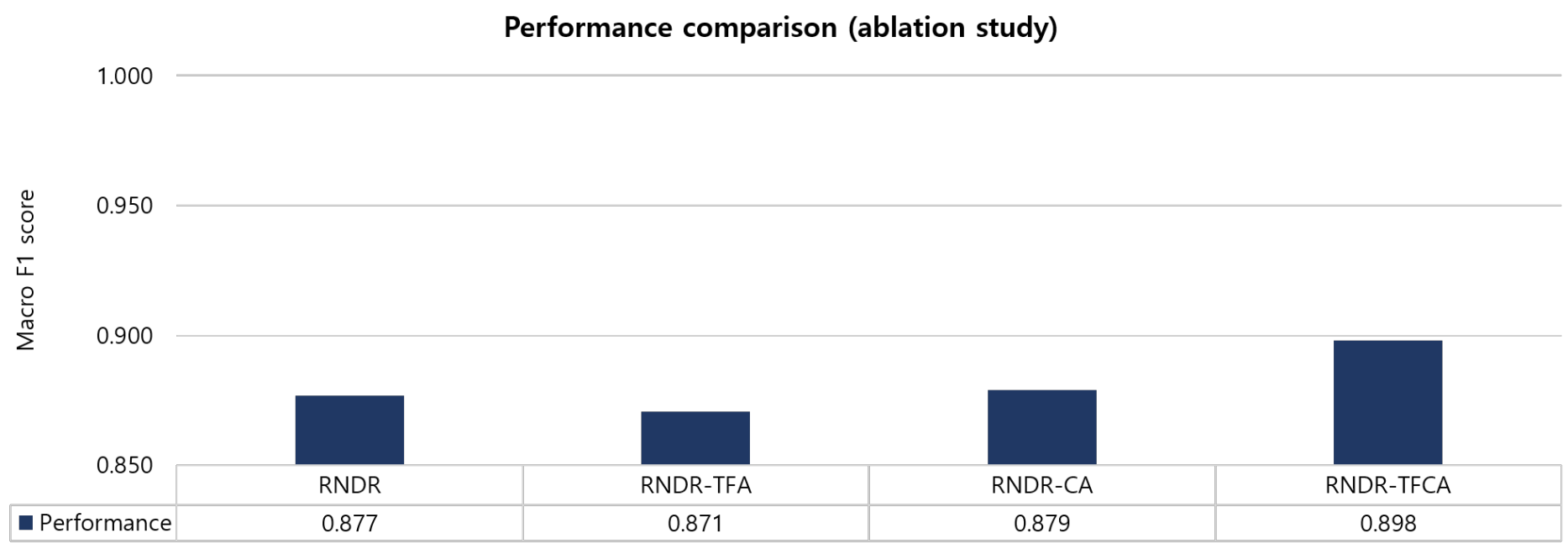

Figure 6: Comparison of Macro F1 scores among ablated variants of RNDR-TFCA.

To investigate the contribution of each attention block, ablation experiments were conducted using RNDR, RNDR-TFA, RNDR-CA, and RNDR-TFCA. RNDR-TFA denotes the variant in which only the time-frequency attention block is applied to enhance intra-task features, whereas RNDR-CA denotes the variant in which only the channel attention block is applied to emphasize inter-task feature components. RNDR-TFCA incorporates both attention blocks, while the original RNDR contains neither.
As shown in Fig. 6, RNDR achieved a Macro F1 score of 0.877, while RNDR-TFA and RNDR-CA achieved 0.871 and 0.879, respectively. Applying either attention mechanism alone therefore resulted in only marginal changes compared with the original RNDR. In contrast, RNDR-TFCA achieved a substantially higher macro F1 score of 0.898. This result indicates that simultaneously enhancing informative intra-task time-frequency regions and inter-task channel-wise feature components is more effective than applying either mechanism independently. The performance gain of RNDR-TFCA therefore suggests a complementary effect between the two attention mechanisms within the CTL.

## 5.3 Model copmlexity and capacity analysis

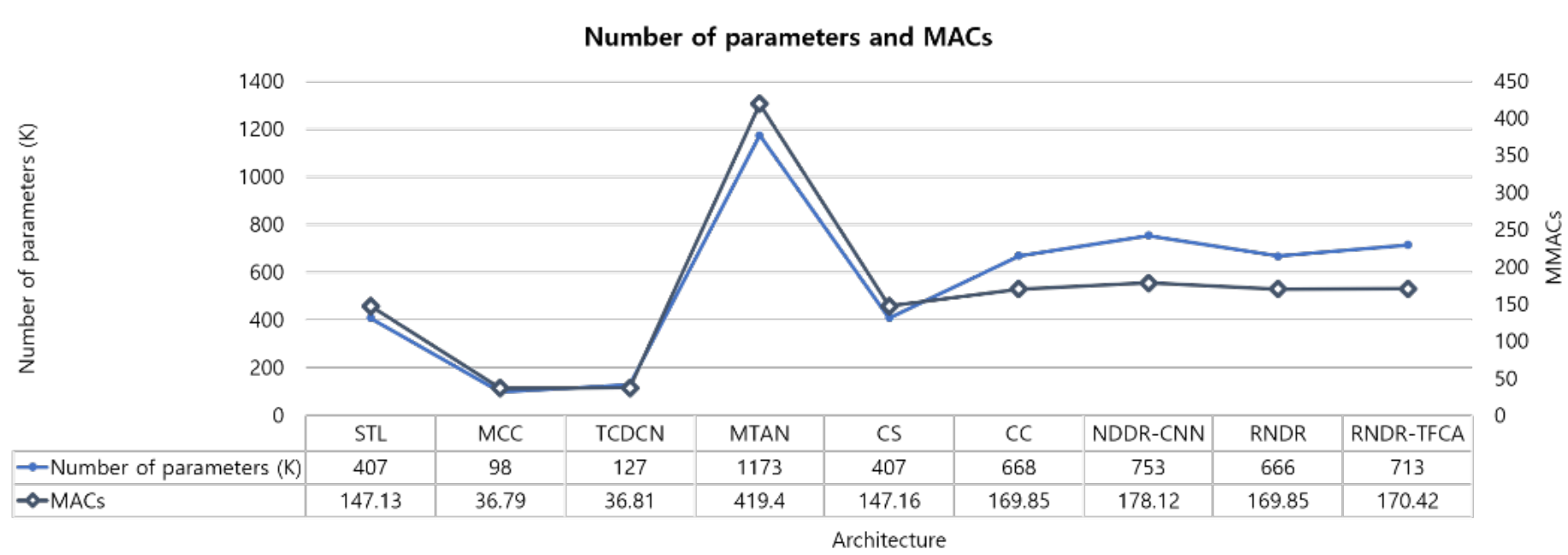

Figure 7: Number of parameters and MACs of STL, MCC, MTL architectures, and proposed RNDR-TFCA.

Fig. 7 compares the number of parameters and MACs of the evaluated architectures. RNDR-TFCA contains 713 K parameters and requires 170.42 MMACs, only slightly higher than RNDR with 666 K parameters and 169.85 MMACs. Its complexity is also comparable to conventional cross-talk architectures, which require 407–753 K parameters and 147.16–178.12 MMACs. These results indicate that the proposed attention modules improve performance without introducing a substantial computational overhead over the RNDR-based architecture.

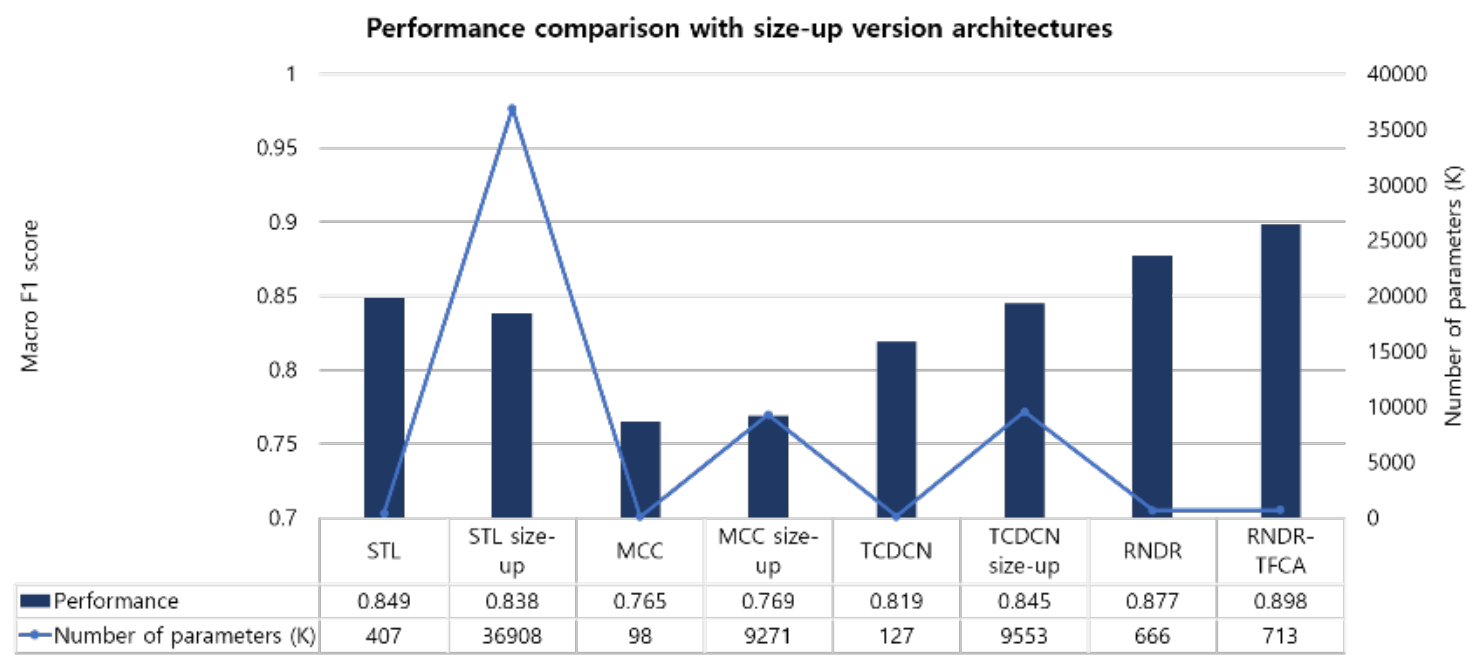

Figure 8: Performance and model-size comparison with channel-expanded baseline architectures.

To assess whether the performance gain was simply due to model capacity, the channel sizes of STL, MCC, and TCDCN were increased tenfold. Despite the substantial parameter increase, their macro F1 scores remained at 0.838, 0.769, and 0.845, respectively, below RNDR (0.877) and RNDR-TFCA (0.898), indicating that the improvement mainly comes from the proposed architecture rather than model size alone.

## 6. Conclusion

This study proposed RNDR-TFCA, an attention-based cross-talk architecture for motor compound fault diagnosis using multi-task learning. RNDR-TFCA extends RNDR by incorporating time-frequency and channel attention into the CTL to enhance intra-task and inter-task information, respectively.
RNDR-TFCA achieved the highest macro F1 score of 0.898, improving the performance of the previously proposed RNDR (0.877) by 0.021, while the conventional methods showed macro F1 scores ranging from 0.765 to 0.876. In the ablation study, RNDR-TFCA achieved a macro F1 score of 0.898, outperforming the original RNDR (0.877) by 0.021 and the single-attention variants, which achieved 0.871–0.879. These results indicate that jointly enhancing intra-task time-frequency information and inter-task channel-wise information is more effective than applying either attention mechanism independently, demonstrating the effectiveness of the proposed attention-based cross-talk architecture for motor compound fault diagnosis. In addition, RNDR-TFCA introduced only limited computational overhead over RNDR, while the channel-expanded baselines still underperformed the proposed method, indicating that the performance gain was not simply attributable to increased model capacity.

## ACKNOWLEDGEMENTS

This work was supported by National IT Industry Promotion Agency(NIPA) grant funded by the Korea government(MSIT) (Development of an Integrated Sim-to-Real Platform for Physical AI-Enabled Collaborative Intelligence, RS-2026-25621650). This work was also supported by the InnoCORE program of the Ministry of Science and ICT (N10260002).

## REFERENCES

1 Gawde, S. et al. Multi-fault diagnosis of industrial rotating machines using data-driven approach: A review of two decades of research. *Engineering Applications of Artificial Intelligence* 123, 106139 (2023).

2 Gawde, S. et al. A scoping review on multi-fault diagnosis of industrial rotating machines using multi-sensor data fusion. *Artificial Intelligence Review* 56, 4711–4764 (2023).

3 Caruana, R. Multitask learning. *Machine Learning* 28, 41–75 (1997).

4 Jia, B.-B. & Zhang, M.-L. Multi-dimensional classification: Paradigm, algorithms and beyond. *Vicinagearth* 1, 3 (2024).

5 Zhang, Z., Luo, P., Loy, C. C. & Tang, X. Facial landmark detection by deep multi-task learning. in *Computer Vision – ECCV 2014* 94–108 (Springer, 2014).

6 Liu, S., Johns, E. & Davison, A. J. End-to-end multi-task learning with attention. in *Proceedings of the IEEE/CVF Conference on Computer Vision and Pattern Recognition* 1871–1880 (2019).

7 Standley, T., Zamir, A., Chen, D., Guibas, L., Malik, J. & Savarese, S. Which tasks should be learned together in multi-task learning? in *Proceedings of the International Conference on Machine Learning (ICML)* 9120–9132 (2020).

8 Misra, I., Shrivastava, A., Gupta, A. & Hebert, M. Cross-stitch networks for multi-task learning. in *Proceedings of the IEEE Conference on Computer Vision and Pattern Recognition (CVPR)* 3994–4003 (2016).

9 Kawakami, R. et al. Cross-connected networks for multi-task learning of detection and segmentation. in *Proceedings of the IEEE International Conference on Image Processing (ICIP)* 3636–3640 (2019).

10 Gao, Y., Ma, J., Zhao, M., Liu, W. & Yuille, A. L. NDDR-CNN: Layerwise feature fusing in multi-task CNNs by neural discriminative dimensionality reduction. in *Proceedings of the IEEE/CVF Conference on Computer Vision and Pattern Recognition (CVPR)* 3205–3214 (2019).

11 Yi, W., Jung, W., Nam, H., Jang, K. & Park, Y.-H. Multi-output classification using a cross-talk architecture for compound fault diagnosis of motors in partially labeled condition. *Mechanical Systems and Signal Processing* 244, 113786 (2026).

12 Lopes, I., Vu, T.-H. & de Charette, R. Cross-task attention mechanism for dense multi-task learning. in *Proceedings of the IEEE/CVF Winter Conference on Applications of Computer Vision (WACV)* (2023).

13 Sinodinos, D. & Armanfard, N. Cross-task affinity learning for multitask dense scene predictions. in *2025 IEEE/CVF Winter Conference on Applications of Computer Vision (WACV)* 1546–1555 (2025).

14 Hu, J., Shen, L. & Sun, G. Squeeze-and-excitation networks. in *Proceedings of the IEEE Conference on Computer Vision and Pattern Recognition (CVPR)* 7132–7141 (2018).

15 Kingma, D. P. & Ba, J. Adam: A method for stochastic optimization. *arXiv preprint arXiv:1412.6980* (2014).

16 Szegedy, C., Vanhoucke, V., Ioffe, S., Shlens, J. & Wojna, Z. Rethinking the inception architecture for computer vision. in Proceedings of the IEEE Conference on Computer Vision and Pattern Recognition (CVPR) 2818–2826 (2016).